\documentclass[conference]{IEEEtran}
\IEEEoverridecommandlockouts
\usepackage{multirow}
\usepackage{booktabs}
\usepackage{cite}
\usepackage{amsmath,amssymb,amsfonts}
\usepackage{graphicx}
\usepackage{textcomp}
\usepackage{xcolor}
\def\BibTeX{{\rm B\kern-.05em{\sc i\kern-.025em b}\kern-.08em
    T\kern-.1667em\lower.7ex\hbox{E}\kern-.125emX}}
    \usepackage{comment}  
\usepackage{texdef2015}
\usepackage{algorithm}
\usepackage{subcaption}
\usepackage[noend]{algpseudocode}
\let\labelindent\relax
\usepackage{enumitem}
\usepackage{setspace}

\newcommand{\age}{\Delta}
\def\BibTeX{{\rm B\kern-.05em{\sc i\kern-.025em b}\kern-.08em
		T\kern-.1667em\lower.7ex\hbox{E}\kern-.125emX}}

\begin{document}


\title{A Deep Learning Approach To Estimation Using Measurements Received Over a Network}


\author{\IEEEauthorblockN{Shivangi Agarwal}
\IEEEauthorblockA{IIIT-Delhi, India\\shivangia@iiitd.ac.in}
\and
\IEEEauthorblockN{Sanjit K. Kaul}
\IEEEauthorblockA{IIIT-Delhi, India\\skkaul@iiitd.ac.in}
\and
\IEEEauthorblockN{Saket Anand}
\IEEEauthorblockA{IIIT-Delhi, India\\anands@iiitd.ac.in}
\and
\IEEEauthorblockN{P. B. Sujit}
\IEEEauthorblockA{IISER Bhopal, India\\sujit@iiserb.ac.in}
}

\maketitle

\begin{abstract}
We propose a novel deep neural network (DNN) based approximation architecture to learn estimates of measurements. We detail an algorithm that enables training of the DNN. The DNN estimator only uses measurements, if and when they are received over a communication network. The measurements are communicated over a network as packets, at a rate unknown to the estimator. Packets may suffer drops and need retransmission. They may suffer waiting delays as they traverse a network path. 

Works on estimation often assume knowledge of the dynamic model of the measured system, which may not be available in practice. The DNN estimator doesn't assume knowledge of the dynamic system model or the communication network. It doesn't require a history of measurements, often used by other works. 

The DNN estimator results in significantly smaller average estimation error than the commonly used Time-varying Kalman Filter and the Unscented Kalman Filter, in simulations of linear and nonlinear dynamic systems. The DNN need not be trained separately for different communications network settings. It is robust to errors in estimation of network delays that occur due to imperfect time synchronization between the measurement source and the estimator. Last but not the least, our simulations shed light on the rate of updates that result in low estimation error.
\end{abstract}
\section{Introduction}
Cyber-physical systems, for example vehicle-to-everything (V2X) and public safety, are amongst the key use cases of next generation networks~\cite{5_6G_usecases}. Such systems have one or more agents actuate in their local physical environment while requiring measurements of their larger environment. For example, a vehicle requires the positions, velocities, and accelerations of vehicles in its vicinity and not sensed by its own sensors. Measurements are communicated as packets, over a communication network, to an agent.

Networks, broadly speaking, introduce packet drops, often resulting in retransmissions, and delays that result from a packet waiting in one or more network queues along a network path, shared by other traffic. Also, network conditions vary over time, due to time-varying traffic and link conditions along any network path. The sources of measurements too choose rates of sending that may be constrained by energy budgets, network connectivity, and sensing rates. 

As a result of the above stated, measurement packets arrive intermittently at an agent and are aged (were generated by a source in the past) when received. To compensate, an agent calculates estimates of measurements at its decision instants, using the history of received packets.

While the problem of estimation has been studied extensively~\cite{hespanha-2007, Schenato_2007_Foundations, park-2017}, the knowledge of the dynamic model of the environment or system being measured is typically assumed. Assuming availability of such model information may not be practical. Say, for example, models for a vehicle's (or driver's) choice of maneuver~\cite{hubmann-2018}. Statistical information, for example the packet drop probability, and a model for the source generating the measurements, is also often assumed. These change over time and are unavailable in practice (see, for example~\cite{yoo-2017-unknown-model}) to the agent calculating estimates. On the other hand, the increase in compute available at edge devices (agents) makes learning based approaches feasible in practice.

We investigate a model-free data driven approach to estimation, which will execute on an agent. Our proposed estimator is a deep neural network. It assumes no knowledge of the measurement source or the dynamic system that is measured or the communications network. Our contributions are next.
\begin{enumerate}[wide, labelwidth=0pt, labelindent=0pt]
    \item We propose a novel Deep Neural Network based approximation architecture, abbreviated as LAA, for the \emph{estimator}, which uses the LSTM cell together with fully-connected layers. LAA takes as input the last estimate, the most recently received measurement and its age at the estimator, circumventing the need to pick a heuristic length of history of measurements received over the network.
    \item We propose a learning algorithm that uses the experience replay buffer, which is commonly used in Deep Reinforcement Learning, in our problem setting. This enables us to train the proposed architecture using measurements received over the network, which may be time correlated.
    \item We demonstrate that LAA doesn't need to be trained separately for specific network instances and may be trained over a generic network with time-varying network parameters. This enables its use in the real-world where network settings may not be known upfront. 
    \item We show that LAA outperforms the Time Varying Kalman Filter (TVKF) and the Unscented Kalman Filter (UKF), respectively, for simulated linear and nonlinear systems, whose measurements are sent to the estimator over a network.
    \item Lack of time synchronisation between an agent and a source of measurements is common in practice. This results in noisy estimates of age of received measurements because of corresponding noisy estimates of network delays. LAA's estimates are robust to noisy estimates.
    \item Simulations show that an age optimal rate of updates results in small estimation errors for all evaluated estimators.
\end{enumerate}

The rest of the paper is organized as follows. We will discuss related works in Section~\ref{sec:related}. This is followed by a description of the system model in Section~\ref{sec:model}. Section~\ref{sec:approxArch} details our proposed LSTM based deep neural network architecture. This is followed by the learning algorithm in Section~\ref{sec:learningAlgo}. Section~\ref{sec:simulatedEnv} describes the varied simulated environments and communication networks used for training and testing the proposed architecture. In Section~\ref{sec:evaluation}, we provide a detailed evaluation of the models we trained using the learning algorithm. We conclude in Section~\ref{sec:conclusions}.

\section{Related Works} 
\label{sec:related}
There is a large body of works under the umbrella of deep reinforcement learning~\cite{deepQNetwork} (DRL) that learns policies for agents in a model-free setting in a data driven manner. Such works~\cite{yadaiah-2006, ayed-2019-dyn_systm_partialobs, meng-2021-memory, ikemoto-2019-cdc19,anagnostopoulos-2021-fingerprintingICC} are relatively limited when the environment is modelled by a time-varying dynamical system and measurements are communicated over an imperfect network. We describe a sampling of such works next.


In~\cite{meng-2021-memory}, the authors propose to use LSTM together with the twin delayed deep deterministic policy gradient algorithm to learn a control policy when measurements and not the state are available. They use a history of measurements and control inputs as an input to their neural network. They don't assume a communications network.  In~\cite{ikemoto-2019-cdc19}, the authors allow for delayed measurements. They assume a known maximum delay and use this to choose the length of history of measurements and controls that are input to a Deep Q Neural network~\cite{deepQNetwork}.

Authors in~\cite{anagnostopoulos-2021-fingerprintingICC} and~\cite{lemic-2020-fingerprinting} use artificial neural networks to estimate localisation error in fingerprinting methods, which are a common solution for indoor positioning systems. In \cite{al-2019-attitude}, a deep neural network is used along with the Kalman filter (assuming a linearised model for a UAV) to learn measurement noise characteristics, in order to improve state estimates. 

In~\cite{leong-2020-deepRLAutomatica}, the authors propose a Deep-Q network for learning a scheduling policy for wireless sensors over a limited number of available channels. They assume that the channel model that governs whether a packet will be successfully received or not is unknown. Also, they assume knowledge of the dynamic model of the system whose state is being estimated. Similarly, in~\cite{yang-2022-Information_science} too a scheduling policy is learnt for sensors observing multiple dynamical systems. However, each sensor transmits measurements to the estimator as opposed to estimates in~\cite{leong-2020-deepRLAutomatica}, for each observed process. 

Works~\cite{yadaiah-2006}  and~\cite{ayed-2019-dyn_systm_partialobs} learn the underlying system dynamic model using measurements. The relatively recent~\cite{ayed-2019-dyn_systm_partialobs} learns the dynamic system model for a continuous time system using measurements of the state of the system and knowledge of the mapping between a state and the corresponding measurement. The much older work~\cite{yadaiah-2006} uses recurrent neural networks to learn the state evolution function and the function that maps a state to a measurement for a discrete time system. These works don't assume a communications network. Also, unlike our work, they attempt to explicitly learn the state evolution and/or measurement functions.

There are works~\cite{hespanha-2007, Schenato_2007_Foundations, park-2017} within the realm of networked control systems that consider the problem of optimal estimation and control when sensors communicate measurements to estimators over a communications network. These works assume knowledge of the dynamical model. Often assumptions are made regarding the model of packet delay and dropout over the communication network. Key to evaluation later in this paper is~\cite{schenato-2008-optimal}, in which the time-varying Kalman filter with an infinite time buffer is shown to be the optimal estimator for a linear system in the presence of intermittent and aged measurements and known controls. Also~\cite{li-2012-stochastic}, which considers estimation using the unscented Kalman filter in presence of intermittent (but not aged) measurements. 

\section{System Model}
\label{sec:model}
A measurement source generates measurements (in general, vectors) $\vvec{y}(t_k)$ at discrete time instants $t_k$, $k=1,2,\ldots$ of a certain environment (dynamical system) of interest. At any instant, it transmits a measurement with probability $0 < p < 1$, which signifies network resource constraints or energy constraints that may preclude scheduling of a generated measurement. The estimator agent calculates estimate $\hat{\vvec{y}}(t^e_k)$ at $t^e_k$, $k=1, 2, \ldots$. It can only use measurements received at times $t \le t^e_k$ to calculate $\hat{\vvec{y}}(t^e_k)$.

Denote the $i$\textsuperscript{th} element of $\vvec{y}(t)$ by $\vvec{y}[i](t)$. The measurements are received by the estimator over a communications network and are transmitted by the source with probability $p$. As a result, the estimator will have access to \emph{aged} measurements that capture an older time snapshot in the evolution of the measured dynamic system. Let $z$ denote a measurement. For example, $z$ could be $\vvec{y}[i]$. Let $U_t(z)$ be the timestamp (time of generation) of the most recently generated measurement $z$ available at the estimator at time $t$. Define the age of measurement $z$ \emph{at the estimator} at $t$ as
\begin{align}
    \age_{z}(t) = t - U_t(z).
    \label{eqn:age_ut}
\end{align}

To exemplify, suppose the most recent timestamp of $\vvec{y}[i]$ at the estimator, at time $t$, is $t'$. That is amongst all measurements of $\vvec{y}[i]$, which the estimator has received so far over the network, the one with a timestamp closest to $t$ is $\vvec{y}[i](t')$. The age of $\vvec{y}[i]$ is $\age_{\vvec{y}[i]}(t) = t - t'$, $t\ge t'$.

Note that $\age_{z}(t)$ at the estimator increases at unit rate in the absence of a measurement with a more recent timestamp than already at the estimator. Further, it is reset to a smaller value in case a measurement with a more recent than available timestamp is received at $t$. $\age_{z}(t)$ is reset to the time elapsed between the generation of the measurement and it being received at $t$~\cite{yates-2021-age}. 


We now describe the communications network. A measurement received by the estimator would have suffered delays because of queuing and (re) transmission. The delays are a function of the rate at which the network can deliver measurements and also the rate at which measurements are sent by the source. To capture the above, we simulate the network as a discrete-time queue with a very large buffer\footnote{An arrival always enters the queue and exits the server post service.} and a single server. The probability $p$ is the arrival rate of measurement packets into the queue. Measurements that enter the queue are serviced in a first-come-first-served manner and each packet spends a geometric$(q)$ time in the server, independently of other updates. That is a packet is communicated in error with probability $1-q$ and re-transmitted till it is successfully received by the estimator. Also, a packet must wait for packets that are ahead of it in the queue to finish service. Note that $p/q$ is the utilization of the network by the source's packets. 

\section{Approximation Architecture}
\label{sec:approxArch}
We propose a Long Short Term Memory (LSTM)~\cite{LSTM} based approximation architecture (LAA), which is a Deep Neural Network, to calculate estimates at $t^e_k$, $k\in \{1,2,\ldots\}$. An LSTM is a sequence model that maps a time sequence of inputs to an output sequence and has the ability to capture long range time dependencies of an output on an input sequence. 

\begin{figure}[tb]
    \centering
    \includegraphics[width =1\linewidth]{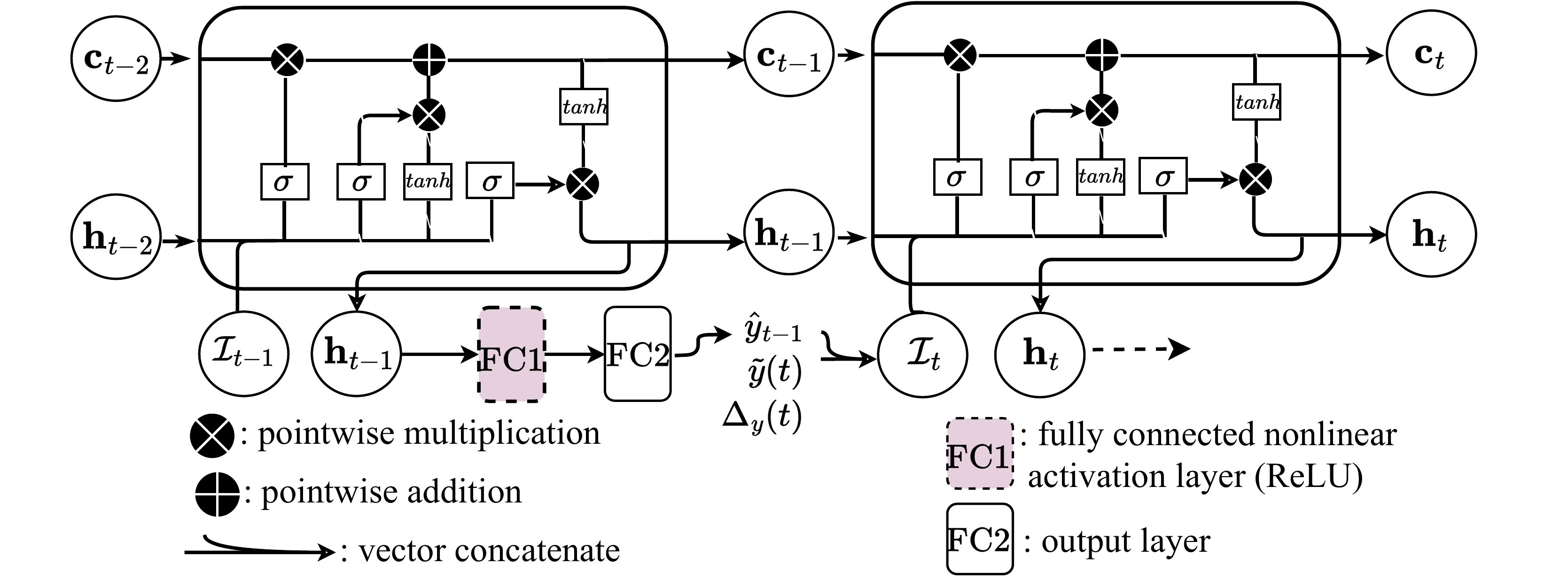}
    \caption{\small LAA unrolled over two time steps. Note that the output $\vvec{h}_{t-1}$ of the LSTM cell (large rectangle with rounded corners) is further processed by two fully-connected layer to yield the estimate $\vvec{\hat{y}_{t-1}}$, which becomes a part of the input at the next time step.}
    \label{fig:LSTM_architecture-unfolded}
    \vspace{-0.2in}
\end{figure}

In our architecture, illustrated in Figure~\ref{fig:LSTM_architecture-unfolded}, at any $t \in \{t^e_1, t^e_2, \ldots\}$ the LSTM cell takes as input the current estimate $\vvec{\hat{y}}(t - 1)$, the received measurement $\vvec{\tilde{y}}(t)$ with the most recent generation timestamp and the corresponding age vector $\age_{\vvec{y}}(t)$ (tracks age for each element in the measurement vector) at the estimator. The input to the LSTM cell at $t$ is
\begin{align}
{\small
    \mathcal{I}(t) = \tvec{\small \vvec{\hat{y}}(t - 1) & \vvec{\tilde{y}}(t) &\age_{\vvec{y}}(t)}.
}\label{eqn:input_LSTM}
\end{align}
Note that the input $\mathcal{I}(t)$ is independent of the total number of measurements received by the estimator up to time $t$. Unlike prior works, our input is not a function of length of (heuristically) chosen history of measurements. 

We pass the output $\vvec{h}(t)$ (also called the hidden state) of the LSTM cell as input to a fully-connected layer FC1 that is parameterized by the weight matrix $\vvec{W}_{\text{FC1}}$ and the bias vector $\vvec{b}_{\text{FC1}}$. The output of this layer is given by $\vvec{o}_{\text{FC1}}(t) = \text{ReLU}(\vvec{W}_{\text{FC1}} \vvec{h}(t) + \vvec{b}_{\text{FC1}})$, where ReLU$()$~\cite{RELU} is the Rectified Linear Unit, a commonly used nonlinear activation function.

The output $\vvec{o}_{\text{FC1}}(t)$ is input to another fully-connected layer FC2, which is parameterized by the weight matrix $\vvec{W}_{\text{FC2}}$ and the bias vector $\vvec{b}_{\text{FC2}}$. The output of the layer FC2 is the estimate $\vvec{\hat{y}}(t) = \vvec{W}_{\text{FC2}} \vvec{o}_{\text{FC1}}(t) + \vvec{b}_{\text{FC2}}$. This estimate becomes a part of the input to the LSTM cell at $t+1$.

Note that the LSTM cell independently updates its \emph{cell state}, its \emph{hidden state}, the \emph{input, forget, cell and output gates}, at every time $t$. Further, the cell contains the \emph{input-hidden} and the \emph{hidden-hidden} weights and biases that must be learnt together with $\vvec{W}_{\text{FC1}}$, $\vvec{b}_{\text{FC1}}$, $\vvec{W}_{\text{FC2}}$, and $\vvec{b}_{\text{FC2}}$. LSTM specific details can be found in~\cite{LSTM}. Figure~\ref{fig:LSTM_architecture-unfolded} illustrates our architecture. In this paper, we set the hidden layer of the LSTM and FC1 to each have $64$ neurons, where the size of a layer is the size of output of the layer. The size of the input layer (vector $\mathcal{I}(t)$) and that of the output layer (FC2) is governed by the dynamic system of interest. For the linear system that we describe later, the size of the input layer is $12$ and that of the output layer is $4$. For the nonlinear system the sizes are $9$ and $3$ respectively.

Let the vector $\vvec{\Theta}$ comprise of all the weights and biases of our proposed architecture. Define the measurement residual as $\vvec{\epsilon}_\vvec{y}(\vvec{\Theta}) = \vvec{y} - \vvec{\hat{y}}(\vvec{\Theta})$. The estimator would like to minimize the mean squared measurement residual~\cite{OptimalControl-Stengel}, with respect to $\vvec{\Theta}$, using measurements obtained over a certain time horizon $T$. Our optimization problem is
\begin{align}
    \underset{\vvec{\Theta}}{\text{minimize}}\ \frac{1}{T} \sum_{t=1}^T \tvec{\epsilon_{\vvec{y}(t)}(\vvec{\Theta})} \vvec{\epsilon}_{\vvec{y}(t)}(\vvec{\Theta}).
    \label{eqn:minimization}
\end{align}
\section{Learning Algorithm}
\label{sec:learningAlgo}
\begin{algorithm}[t]
    \caption{Learning Algorithm For LAA}
      \label{alg:learning-LAA}
      \footnotesize
      \begin{algorithmic}[1]
              \State{Initialize weights $\vvec{\Theta}$ and replay memory $\mathcal{D}$.}
              \For{episode = $1$, $M$}
                \State{Initialize dynamic system (environment); System state $\vvec{x}(1)$ is set.}\label{alg:initEpisode}
                \State{Initialize network; set $p,q$.}
                \State{At the estimator: Initialize $\vvec{\hat{y}}(0)$, $\vvec{\tilde{y}}(1)$, $\age_{\vvec{y}}(1)$.}
                \For{$t = 1, T$}
                
                    \textbf{Dynamic System and Network Simulation:}\label{alg:systemSim}
                    \State{Source generates $\vvec{y}(t)$, a measurement of the environment.}\label{alg:generateMeas}
                    \State{With prob. $p$ queue a packet containing $\vvec{y}(t)$.}
                    \If{packet received at the estimator }~\label{ageUpdateStart}
                        \State{Update $\vvec{\tilde{y}}(t)$; reset $\age_{\vvec{y}}(t)$ as in~(\ref{eqn:age_ut}).}
                    \Else
                        \State{$\vvec{\tilde{y}}(t)\leftarrow\vvec{\tilde{y}}(t-1)
                        $, $\age_{\vvec{y}}(t)\leftarrow\age_{\vvec{y}}(t-1)+1$.}
                    \EndIf~\label{ageUpdateEnd}
                    
                    \State{Dynamic system transitions to state $\vvec{x}(t+1)$.}\label{alg:systemSimEnd}
                    
                    \textbf{Estimator:}
                    
                    \State{Store experience $\{\mathcal{I}(t), \vvec{{y}}(t)\}$ in $\mathcal{D}$.}\label{alg:estimator}\label{alg:storeinReplay}
                    \State{Sample mini-batch containing $K$ transitions from $\mathcal{D}$ randomly.}\label{alg:fwdPassBegin}
                        \For{Each experience in mini-batch.}
                        \State{\emph{Forward Propagation:} Calculate $\vvec{\hat{y}}$ using stored $\mathcal{I}$ and $\vvec{\Theta}$.}
                        \State{Calculate squared measurement residual $\tvec{\epsilon_{\vvec{y}}(\vvec{\Theta})}\vvec{\epsilon}_{\vvec{y}}(\vvec{\Theta})$.}
                        \EndFor\label{alg:fwdPassEnd}%
                    \State{\emph{Backpropagation:} Update $\vvec{\Theta}$ by performing a gradient descent step on the squared residual averaged over the mini-batch.}\label{alg:backProp}
                \EndFor
            \EndFor
      \end{algorithmic}
\end{algorithm}
Algorithm~\ref{alg:learning-LAA} summarizes the learning algorithm that is used to train the LAA model (solve problem~(\ref{eqn:minimization})) using data generated by simulating the dynamic system (environment). The algorithm is executed at the estimator, which updates the weight vector $\vvec{\Theta}$ at time instants $t_k^e$, using measurements it has received up to then. 


The calculation of $\vvec{y}(t)$ and transmission of packets over the communications network, shown in Lines~\ref{alg:systemSim} --~\ref{alg:systemSimEnd} of the algorithm, captures simulation of the network and generation of measurements. The processing at the estimator, starting Line~\ref{alg:estimator}, captures the training of LAA. Having the two execute in sequence simplifies the learning setup and presentation. In general, they execute independently of each other. 

In our experiments, the training of LAA is executed over $M\ge 200$ episodes with each episode $T = 40000$ time slots long. The algorithm begins by the estimator initializing the weight vector $\vvec{\Theta}$ and the replay memory (explained later) $\mathcal{D}$. At the beginning of every episode (Line~\ref{alg:initEpisode}), the dynamic system is initialized and as a result its state $\vvec{x}(1)$ is set. The network is initialized with a choice of $p$ and $q$. The estimator initializes the input to LAA.

At any time step $t$ in an episode, the source generates a measurement vector $\vvec{y}(t)$ of the dynamic system (see Line~\ref{alg:generateMeas}). Next, with probability $p$, the measurement is added to the network queue. Any update packet that completes service is received by the estimator. The last received measurement and its age are appropriately updated (Lines~\ref{ageUpdateStart} -- \ref{ageUpdateEnd}). 


The estimator stores its experience (Line~\ref{alg:storeinReplay}) at $t$ in the replay memory $\mathcal{D}$. The experience includes the input $\mathcal{I}(t)$ to LAA and the ground-truth measurement $\vvec{y}(t)$. It must be pointed out that the ground-truth is only needed while the estimator is in the process of learning a good weight vector $\vvec{\Theta}$, that is when LAA is being trained.\footnote{While we assume that the ground-truth is readily available, in practice the said ground-truth is only available if and when the measurement $\vvec{y}(t)$ is received by the estimator over the communications network. Assuming that the latter is the case, the estimator must add the experience corresponding to the input $\mathcal{I}(t)$ if and when $\vvec{y}(t)$ is received. This would result in experiences being added to the replay memory at a slower rate than when the ground-truth is readily available. However, the algorithm stays unchanged.}.

In addition, at every time $t$ (even if no new experience was added, as may happen in practice) the estimator calculates the average squared error over a randomly chosen mini-batch (Lines~\ref{alg:fwdPassBegin} --~\ref{alg:fwdPassEnd})\footnote{Referred to as a mini-batch as it is typically a small subset of all experiences in the replay memory.} of experiences in the replay memory. Note that the calculation of error requires \emph{forward propagation}~\cite{deepQNetwork} using LAA. Specifically, for each experience we must input the corresponding $\mathcal{I}$ to LAA and obtain the $\vvec{\hat{y}}(t)$. The error is then used to update $\vvec{\Theta}$ using gradient descent. This requires calculation of gradient of the mean squared measurement residual cost function with respect to the weight vector $\vvec{\Theta}$ (referred to as \emph{Backpropagation}~\cite{deepQNetwork}, see Line~\ref{alg:backProp}).

\emph{On the replay memory:} The replay memory is often used to ensure that consecutive weight updates result from experiences that are independent of each other. Choosing experiences in the time sequence in which they occur would result in strongly correlated experiences when updating the weight vector, which is detrimental to learning. The use of replay memory, when training a deep neural network, to break correlations is common (for example, see~\cite{deepQNetwork}).

\emph{On Hyperparameters and computational complexity:} We used the Adam optimiser with learning parameter $10^{-4}$ and mini-batch size of $256$. To prevent the model from overfitting, we use a L$2$ weight decay of $10^{-3}$. Our replay memory is of size $2\times 10^6$ experiences. The LSTM cell has four gates, each of which computes two matrix-vector multiplications and two additions involving vectors. For an input of size $n_x$ and a LSTM cell with $n_h$ neurons in the hidden layer, the total number of computations per gate is $n_x n_h + n_h + n^2_h + n_h$. Taking into account the cell state and the hidden state calculations in the LSTM cell, and the two fully connected layers FC1 and FC2 (of size $n_h$ and $n_o$ respectively), the total number of computations can be approximated by $4 (n_x n_h + 2n_h + n^2_h) + 4n_h + 2 n_h^2 + n_h n_o$.
\section{Simulated Environments and Networks}
\label{sec:simulatedEnv}
We simulated the two environments of a vehicle moving in two dimensional space and that of an inverted pendulum (a pole) fixed to a moving cart, often referred to as the cartpole. We simulated different network configurations via appropriate choices of $p$ and $q$. We detail both next.
\subsection{Simulated Environments}
The vehicle's motion is well modeled by a linear dynamic system that follows Newtonian kinematics. Such a model for a vehicle is often used in autonomous vehicle path planning (see, for example,~\cite[Equation~($11$)]{hubmann-2018}) to model the motion of other vehicles whose positions and velocities an autonomous vehicle must estimate. The measurement vector of the vehicle at time $t_k$ is $\vvec{y}_{k} = \tvec{{p}_{x,k} & {p}_{y,k} & {v}_{x,k} & {v}_{y,k} & {u}_{x,k} & {u}_{y,k}}$, wherein $(p_{x,k}, p_{y,k})$ are the position coordinates, $(v_{x,k}, v_{y,k})$ are the velocities along the $x$ and $y$ dimensions, and $(u_{x,k}, u_{y,k})$ are the corresponding accelerations. The position coordinates are allowed to vary over $[-1000,1000]$ m and the velocities over $[-10,10]$ m/sec. At every time $t_k$, the accelerations along $x$ and $y$ are chosen uniformly and randomly from $[-3,3]$ m/s$^2$. The vehicle's motion is given by Equation~(\ref{eqn:linearVehDyn}), where the noise vector $\vvec{w_k}$ is a Gaussian noise vector of mean $0$ and a diagonal covariance matrix, with the diagonal elements set to $0.2$. The simulation time step is $\Delta t = t_{k+1} - t_k=0.1$ s, for all $k$.

LAA estimates $\vvec{\hat{y}}_{k} = \tvec{\hat{p}_{x,k} & \hat{p}_{y,k} & \hat{v}_{x,k} & \hat{v}_{y,k}}$. Note that, the vehicle state at any time is summarized by its position and velocity. So while the measurement may include acceleration information and is useful for prediction, the estimated vector has only position and velocity.
\begin{table}[b]
{
\footnotesize
\begin{align}
\hline\nonumber\\
&\smvec{p_{x,{k+1}}\\ p_{y,{k+1}} \\ v_{x,{k+1}} \\ v_{y,{k+1}}} = \smvec{
1 & 0 & \Delta t & 0 \\
0 & 1 & 0 & \Delta t \\
0 & 0 & 1 & 0 \\
0 & 0 & 0 & 1} \smvec{p_{x,{k}}\\ p_{y,{k}} \\ v_{x,{k}} \\ v_{y,{k}}} + \smvec{\frac{(\Delta t)^2}{2} & 0 \\
0 & \frac{(\Delta t)^2}{2} \\
\Delta t & 0 \\
0 & \Delta t} \bmat{u_{x,k} \\
u_{y,k}} + \vvec{w_k}.\label{eqn:linearVehDyn}\\
&\ddot{\theta} = \frac{g\sin\theta+\cos\theta\left(\frac{-F-m_pl\dot{\theta^2}\sin\theta}{m_c+m_p}\right)}{l\left(\frac{4}{3}-\frac{m_p\cos^2\theta}{m_c+m_p}\right)},
\ddot{x} = \frac{F+m_pl(\dot\theta^2\sin\theta-\ddot\theta\cos\theta)}{m_c+m_p}.\label{eqn:cartpole}
\end{align}
}%
\end{table}

The cartpole environment is a nonlinear dynamic system that has a pole attached to a moving cart. Ideally, one wants to move the cart to stabilize the pole (keep it close to the vertical position) attached to it. We simulate the environment as implemented by OpenAI gym api~\cite{brockman-2016openai}. The continuous-time dynamics are given by Equation~\ref{eqn:cartpole}, where $F$ is the force applied to the cart, $m_c$ is mass of the cart, $m_p$ and $2l$ are respectively the mass and length of the pole, $x$ is the position of the cart, $\theta$ is the angle between the pole and the vertical, and $g$ is the acceleration due to gravity. We set $l=1$ m, $m_c=5$ kg and $m_p = 1$ kg. The cart's velocity $\dot x$ stays in the interval $[-10,10]$ m/sec. At every time $t_k$, we randomly choose the direction of motion of the cart to be either forward or reverse with equal probability. The simulation timestep $t_{k+1} - t_k=0.01$ s, for all $k$. 

The measurement vector of the cartpole at time $t_k$ is $\vvec{y}_{k} = \tvec{{\theta}_{k} & {\dot\theta}_{k} & {\dot x}_{k} & F_k}$. LAA would like to estimate the first three quantities in the measurement vector.  

\subsection{Communication Networks} 
\label{sec:communication_networks}

\begin{figure}[t]
    \centering
    \includegraphics[trim=0cm 1cm 0cm 1cm, width = \linewidth,clip]{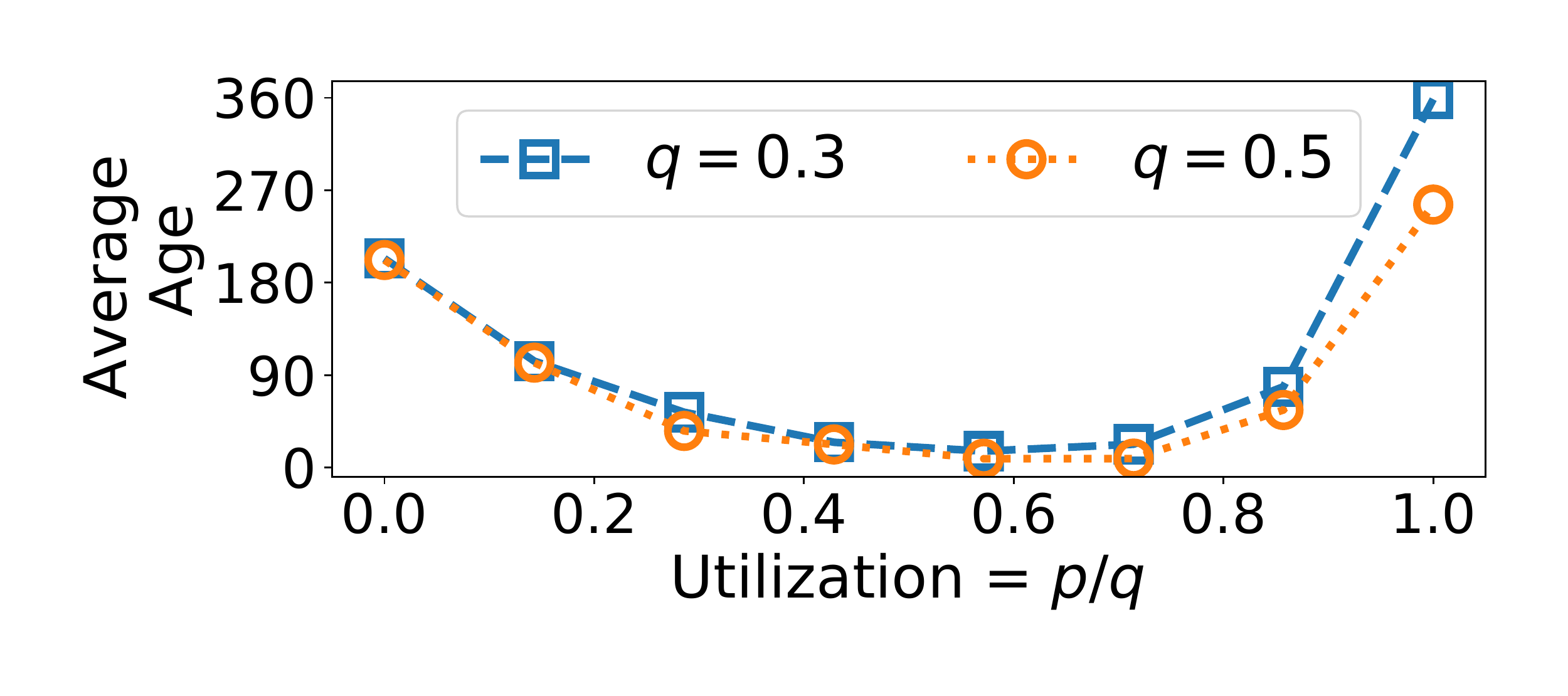}
    \caption{\small Average age as a function of utilization for $q=0.3, 0.5$.}
    \label{fig:averageAgevsp}
    \vspace{-0.1in}
\end{figure}
We consider example networks with fixed $p$ and $q$ for all time and also a network in which $p$ and $q$ change over time. For when the network parameters are time-invariant, we choose $p=0.01, 0.1, 0.297$ for $q=0.3$, and $p=0.01, 0.3, 0.499$ for $q=0.5$. The choices of $p$ for each $q$ are motivated by Figure~\ref{fig:averageAgevsp} that shows how the average age of updates varies at the estimator as a function of $p$ for $q=0.3, 0.5$. As is seen in the figure, the average age is high for both $p=0.01$ and $p=0.297$ when $q=0.3$. As has been shown to hold true for a wide range of network settings~\cite{yates-2021-age}, the high age at $p=0.01$ is explained by a low rate $p$ of sending of updates over the network. A small $p$ with respect to $q$ results in low utilization of the queue server by the update packets. An update that arrives to the queue suffers small queue wait times before completing service. As a result, updates received by the estimator have a small age and are relatively recent. However, the average age of updates at the estimator is still high because the server receives updates infrequently. 

The average age is high at $p=0.297$ because the rate of sending updates is close to $q$. The high utilization $p/q$ of the server has updates experience high delays due to large queue waiting times. The estimator receives updates at a high rate $p$ but with a high average age. The high average ages when $p=0.01, 0.499$ for $q=0.5$ can be explained similarly. 
The network settings of $p=0.1, q=0.3$ and $p=0.3, q=0.5$ result in close to minimum average age.

\section{Evaluation}
\label{sec:evaluation}


\begin{figure}[t]
\begin{center}
\subfloat[]{\includegraphics[width = 0.5\linewidth] {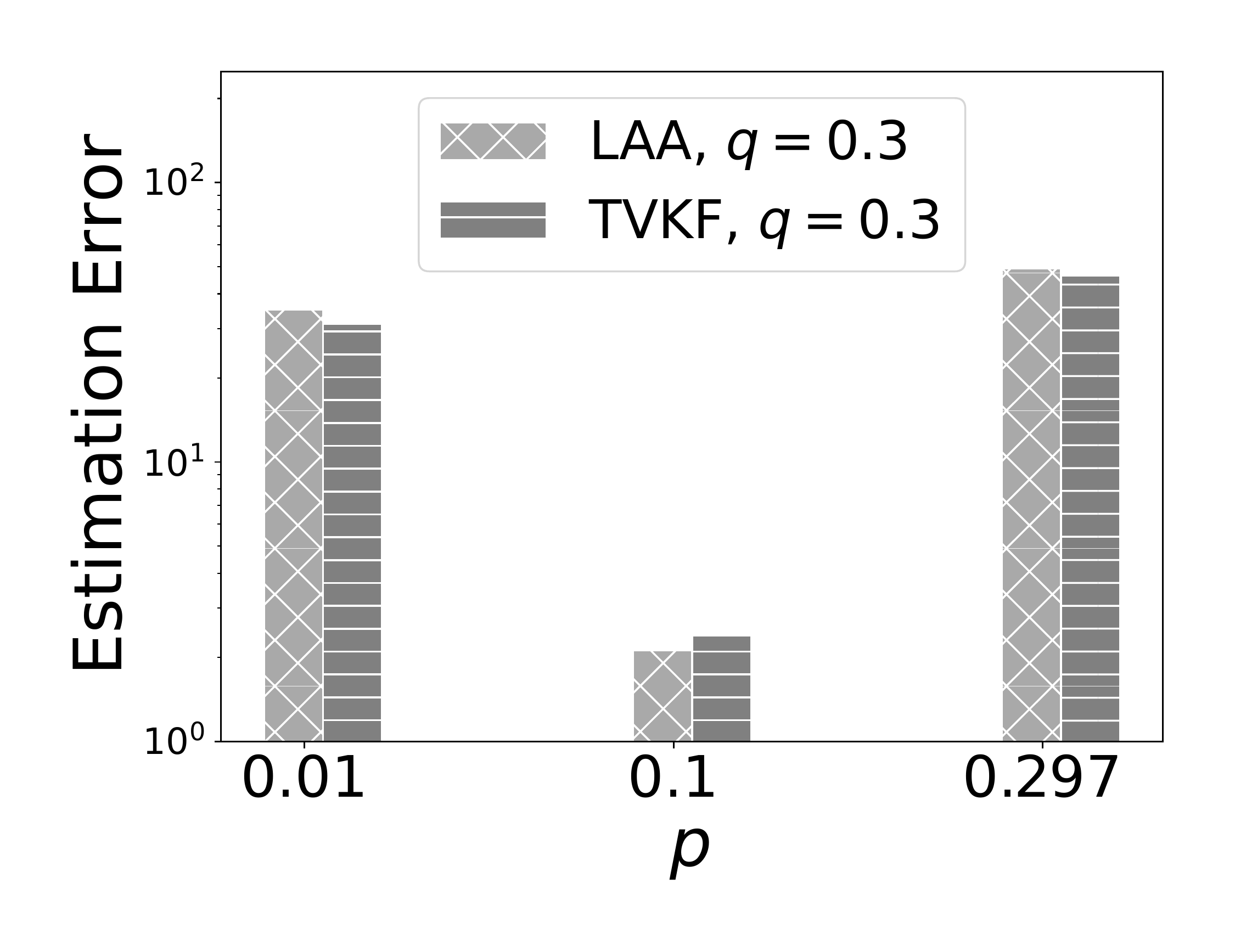}
\label{fig:baselinear_age}}
\subfloat[]{\includegraphics[width = 0.5\linewidth] {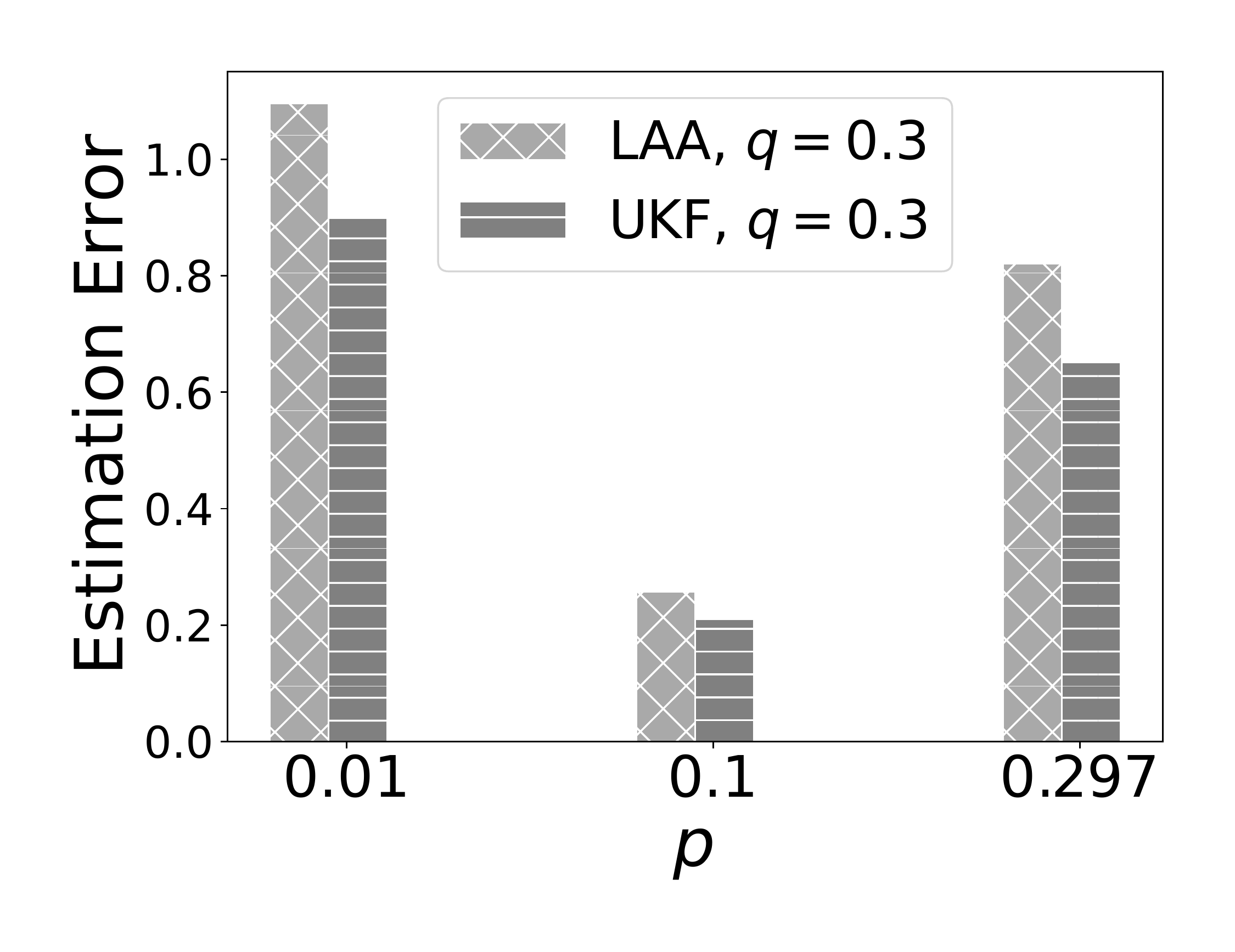}
\label{fig:baseline_nonlinear_age}}
\caption{\small Comparison of LAA and (a) the TVKF and (b) the UKF, for the \emph{known controls} setting.}
\label{fig:baseline_comparisons}
\end{center}
\vspace{-0.2in}
\end{figure}

We used the learning algorithm, described in Algorithm~\ref{alg:learning-LAA} to train a LAA model for each of the following network and dynamic system settings. For both the linear system and the cartpole, we trained a model for each $(p,q)$ in the list $(0.01, 0.3)$, $(0.1, 0.3)$, $(0.297, 0.3)$, $(0.01, 0.5)$, $(0.3, 0.5)$, $(0.499, 0.5)$. We also trained a model, where in each episode of training we randomly chose $q$ in $(10^{-2}, 1)$ and $p$ in $(10^{-3}, q)$
\footnote{We sample uniformly from the range of $log10$ probabilities to ensure a good representation of probabilities $<< 1$.}. Note that it is essential for queue stability to choose $p < q$. We will refer to the former settings as \emph{fixed network} and the latter as \emph{time-varying network}. The two are compared in Section~\ref{sec:general_vs_fixed}.

Further, for each fixed-network setting, we perform an ablation study to answer whether the age $\age_{\vvec{y}}(t)$ at the input of LAA (see Equation~(\ref{eqn:input_LSTM})) makes a significant difference to the estimation error. To do so, we train a LAA model, for each fixed network setting, without age as an input. We will refer to this as the \emph{No Age} setting, which is evaluated in Section~\ref{sec:w/o_age}.

Last but not the least, for each fixed-network setting, we train a LAA model, for when the current values of $u_{x,k}, u_{y_k}$ and $F_k$ are assumed to be known to the estimator, respectively for the linear system and the cartpole. These models help us baseline the performance of LAA, as we explain later in Section~\ref{sub:baseline}. The acceleration values $u_{x,k}, u_{y_k}$ serve as control actions for the linear system. Similarly the force $F$ serves as a control action for the cartpole. This when the control actions are assumed known is referred to as the \emph{known control} setting.

In Section~\ref{sec:noisy_age}, we test the trained models that take age as an input for when the true age is available and also when a noisy estimate of the same is available. We also test the fixed network models that don't take age as input. The performance metric of interest is the estimation error, which is the square-root of the average squared measurement residual defined in Section~\ref{sec:approxArch}. We show a comparison of the estimation error achieved by the LAA models, for the linear dynamic system, with the commonly used Time-Varying Kalman Filter (TVKF), and for the cartpole, with the Unscented Kalman Filter (UKF). The average error is computed over $200$ episodes, wherein each episode is $40000$ time steps long.

\subsection{Baselining LAA using the known control setting}
\label{sub:baseline}
We begin by evaluating LAA for the \emph{known control} setting. For this setting, the baseline algorithms of Time-Varying Kalman Filter (TVKF) and the Unscented Kalman Filter (UKF) for the non-linear dynamic system are expected to do well. Both TVKF and the UKF assume knowledge of the dynamic system model. 

The TVKF with infinite time buffer is in fact known to be the optimal estimator~\cite{schenato-2008-optimal}, for linear dynamic systems, when the controls actions are always known to the estimator. This gives us an opportunity to benchmark the data-driven LAA against a known optimal estimator. We compare (see Figure~\ref{fig:baselinear_age}) the estimation error obtained using the LAA models trained for the \emph{known control} setting with that obtained using the TVKF for the linear system. The algorithms are compared for when $q=0.3$ and $p$ takes values of $0.01$ (low rate and high age, see Figure~\ref{fig:averageAgevsp}), $0.1$ (close to the age optimal rate), and $0.297$ (high rate, delay, and age). \emph{As is seen from the figure, LAA performs almost as well as the optimal TVKF.}

For a nonlinear dynamic system, we were able to find a UKF implementation~\cite{li-2012-stochastic} that works with intermittent measurements. However, an update when received is assumed current (fresh, age $0$). We modify the UKF implementation in the following manner. When a measurement is not received, we propagate the prior state estimate and error covariance according to the system dynamic model and set the posterior of the state and error covariance to their respective priors. When a new measurement arrives, we use it to recalculate the posterior estimates corresponding to the timestamp of the update, and propagate the estimates up to the current time using the model. Figure~\ref{fig:baseline_nonlinear_age}, compares LAA and UKF. \emph{LAA compares well with UKF.}

\begin{figure}[t]
\begin{center}
\subfloat[]{\includegraphics[width = 0.5\linewidth] {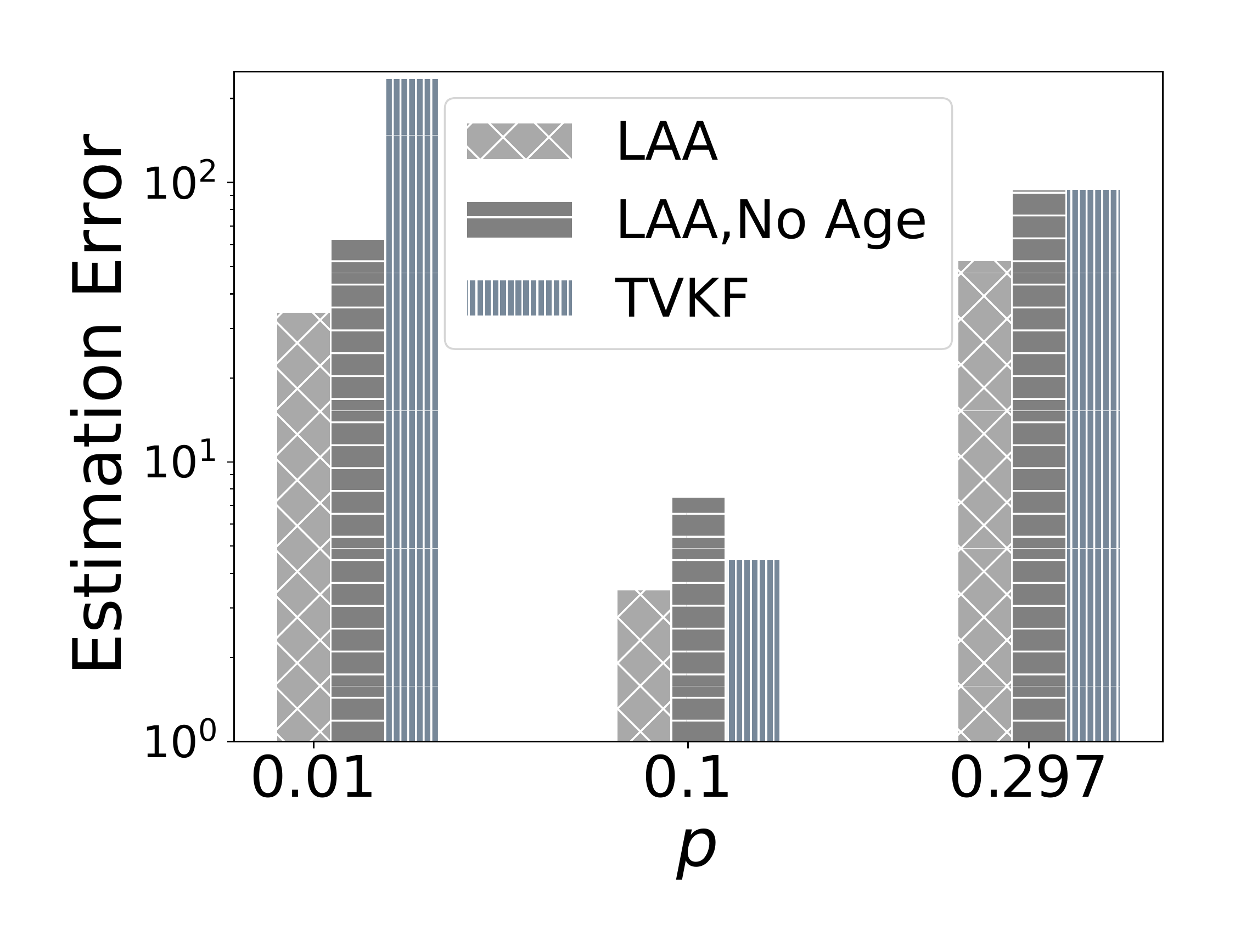}
\label{fig:agevsnoage_0.3_linear}}
\subfloat[]{\includegraphics[width = 0.5\linewidth] {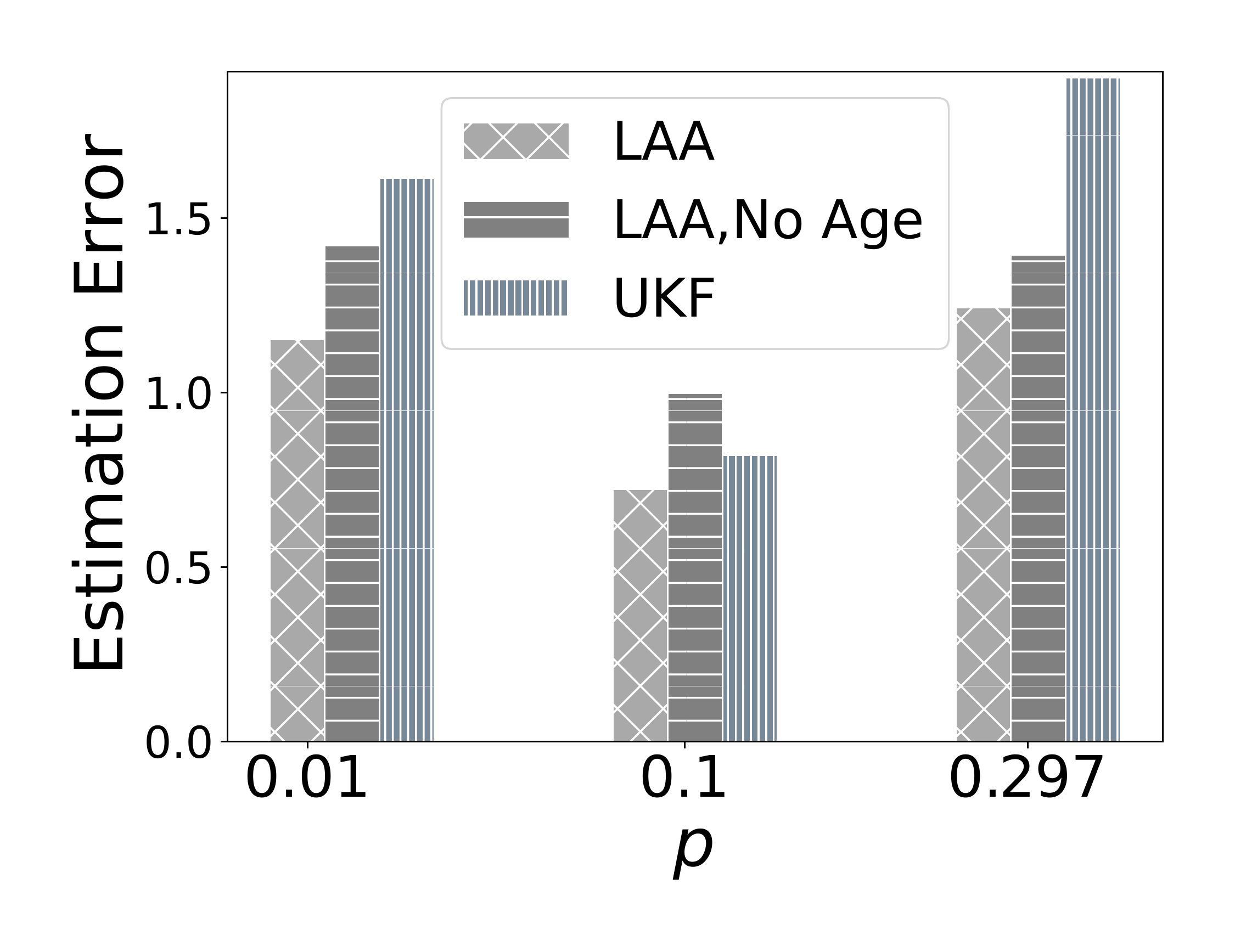}
\label{fig:agevsnoage_0.3_nonlinear}}
\caption{\small Figures summarize the impact of not including age at the input of LAA. Controls are communicated over the network for $q=0.3$. (a) Linear system,  (b) Nonlinear. Results for $q=0.5$ are not shown but look similar. Including age at the input of LAA is always better estimation error wise.}
\label{fig:AgeOrNot}
\end{center}
\vspace{-0.2in}
\end{figure}

\subsection{LAA's performance for the fixed network setting, with and without age inputs}
\label{sec:w/o_age}
As in typically the case, the control actions, unlike for baselining, are obtained over the network. We assume that the TVKF and the UKF use the last known control till a more recent measurement packet containing controls is received over the network. For the linear system, Figure~\ref{fig:agevsnoage_0.3_linear} compares the estimation error obtained by LAA (\emph{fixed network} setting), LAA with \emph{No Age}, and the TVKF. Figure~\ref{fig:agevsnoage_0.3_nonlinear} does the same for the cartpole with the UKF chosen instead of the TVKF. It is clear that LAA with the age inputs does the best and in fact outperforms the TVKF and the UKF. LAA without the age inputs performs fairly well when $p$ is set close to the age optimal rate but suffers otherwise. \emph{Keeping the age inputs $\vvec{\age_u}$ and $\vvec{\age_y}$ clearly benefits estimation. Also, see Figures~\ref{fig:baseline_comparisons} and~\ref{fig:AgeOrNot}, all estimators give a much smaller estimation error when the chosen $p$ is the age optimal rate.}


\subsection{LAA's performance when age estimates are noisy}
\label{sec:noisy_age}
In practice, getting an accurate estimate of age requires the source of the measurement and the estimator to be time-synchronized. This is often imperfect and typically not available. A typical workaround is to estimate one-way delays using round-trip-time (RTT) (between source and estimator) measurements. The one-way delay estimates can then be used to estimate age, which recall is the time elapsed between generation of a measurement and its reception at the estimator.

\begin{figure}[t]
	\begin{center}
		\subfloat[]{\includegraphics[width = 0.5\linewidth]{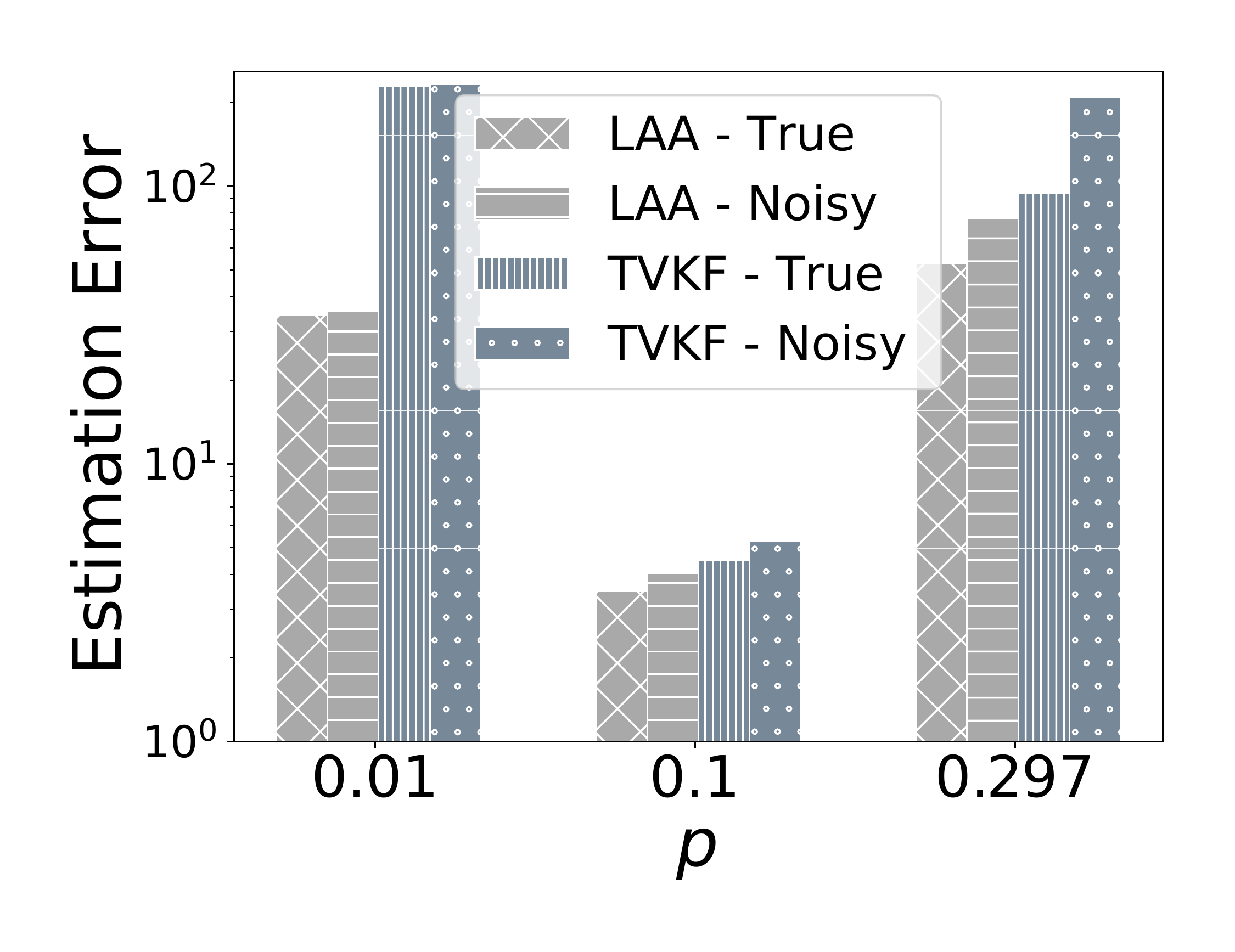}
			\label{fig:noisy_linear_0.3}}
		\subfloat[]{\includegraphics[width = 0.5\linewidth]{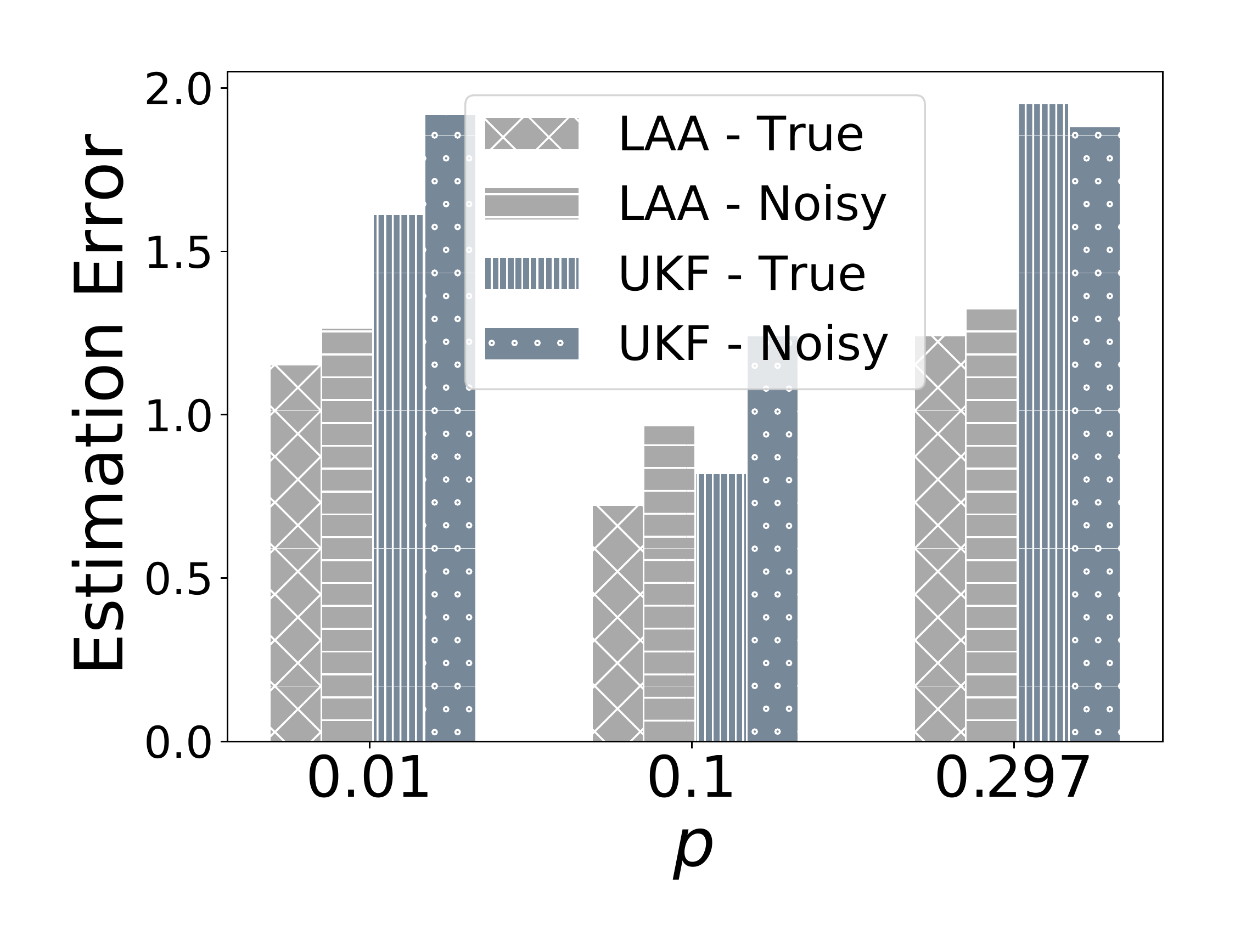}
			\label{fig:noisy_nonlinear_0.3}}
		\caption{\small Impact of noisy age estimates (a) Linear $q=0.3$ (b) Nonlinear $q=0.3$. Controls are communicated over the network. The estimation error isn't impacted by noisy age inputs.}	
    	\label{fig:noisyAgeEst}		
	\end{center}
	\vspace{-0.2in}
\end{figure}

For the purpose of evaluating the impact of noisy age estimates on LAA's performance, we scale the true age by a continuous uniform random variable that takes values in $(0,2)$, where a scaling of $2$ approximates a RTT based estimate of age. To the scaled value we add a Gaussian with mean $0$ and standard deviation equal to $10\%$ of the true age. Figure~\ref{fig:noisyAgeEst} shows the impact of noisy age estimates on the estimation error. Note that LAA was trained as earlier with true age values at its input. Noisy age estimates were only provided when testing. \emph{The estimation performance of LAA is not very impacted, unlike the TVKF and the UKF that see a significant increase in estimation error in comparison to when the true age is available.}
 

\subsection{Does LAA train well over a time-varying network setting? Or do we need to train it separately for every fixed network setting of interest?}
\label{sec:general_vs_fixed}
\begin{figure}[t]
	\begin{center}
		\subfloat[]{\includegraphics[width = 0.5\linewidth]{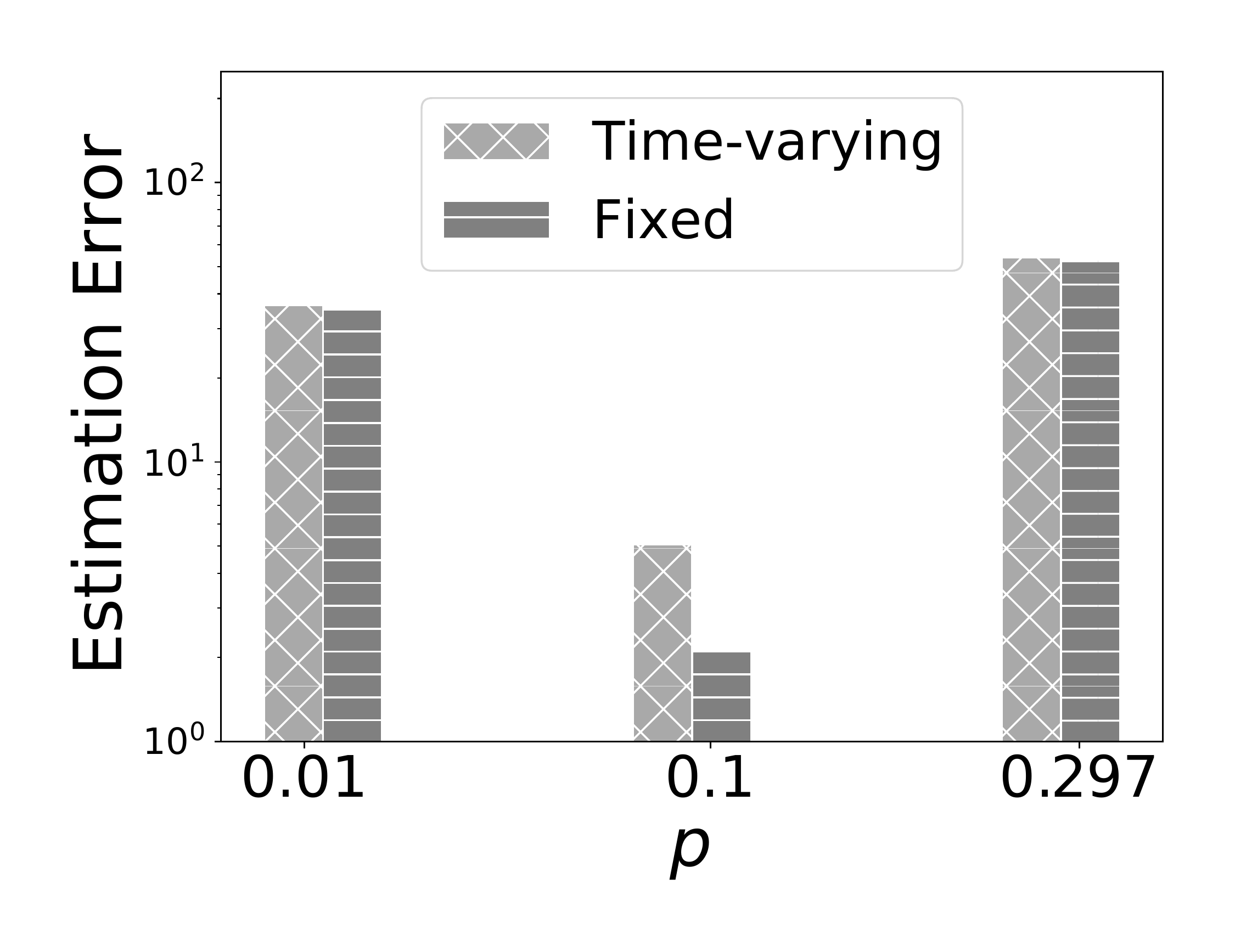}
			\label{fig:generalvsfix_linear_0.3}}
		\subfloat[]{\includegraphics[width = 0.5\linewidth]
		{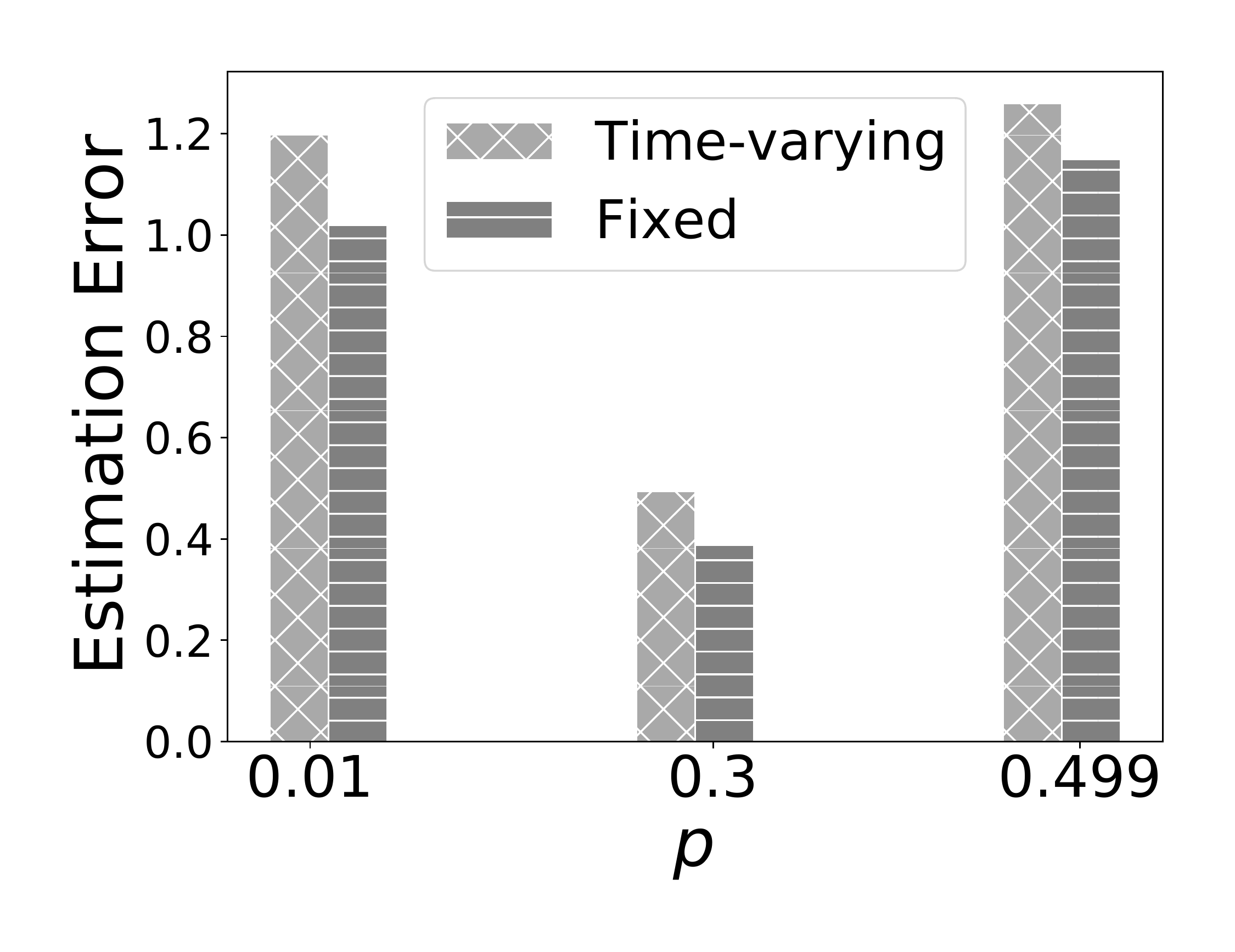}
			\label{fig:generalvsfix_nonlinear_0.5}}
		\caption{\small Figures compare the test performance of the LAA model trained over a time-varying network setting with the performance of a model trained for the specific network setting used for the test. Controls are communicated over the network. (a) Linear system, $q=0.3$ and the $p$ shown in the figure are the different test network settings (b) Nonlinear, $q=0.5$ and the $p$ shown in the figure are the different test network settings. The corresponding results for $q=0.5$ and $q=0.3$ are similar. The LAA model trained over a time-varying network does almost as well as the models trained for specific network settings.}	
    	\label{fig:TimeVaryingNtwkTrain}		
	\end{center}
	\vspace{-0.125in}
\end{figure}

\begin{figure}[t]
	\begin{center}
		\subfloat[]{\includegraphics[width = 0.5\linewidth]{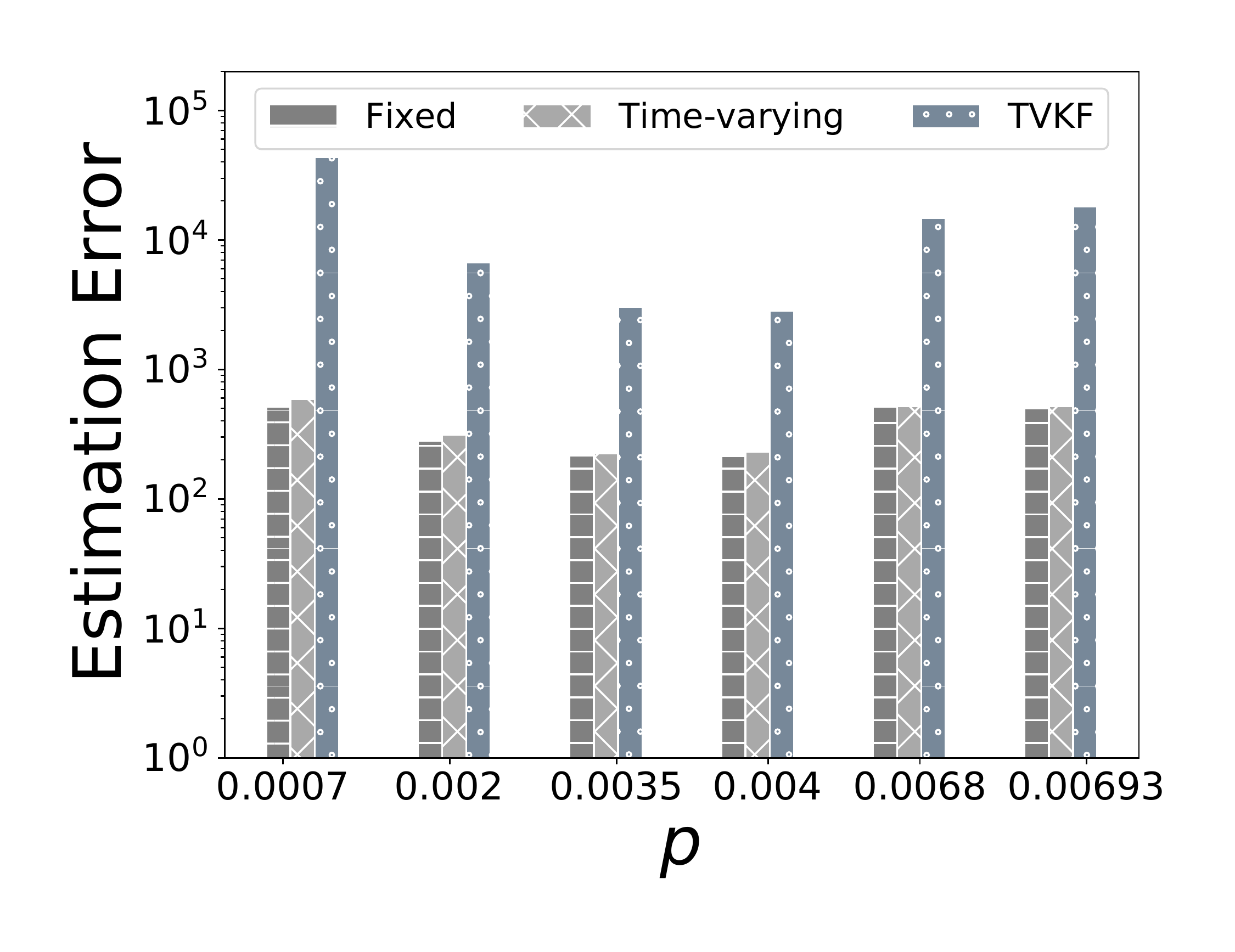}
			\label{fig:generalvsfix_linear_0.0007}}
		\subfloat[]{\includegraphics[width = 0.5\linewidth]
		{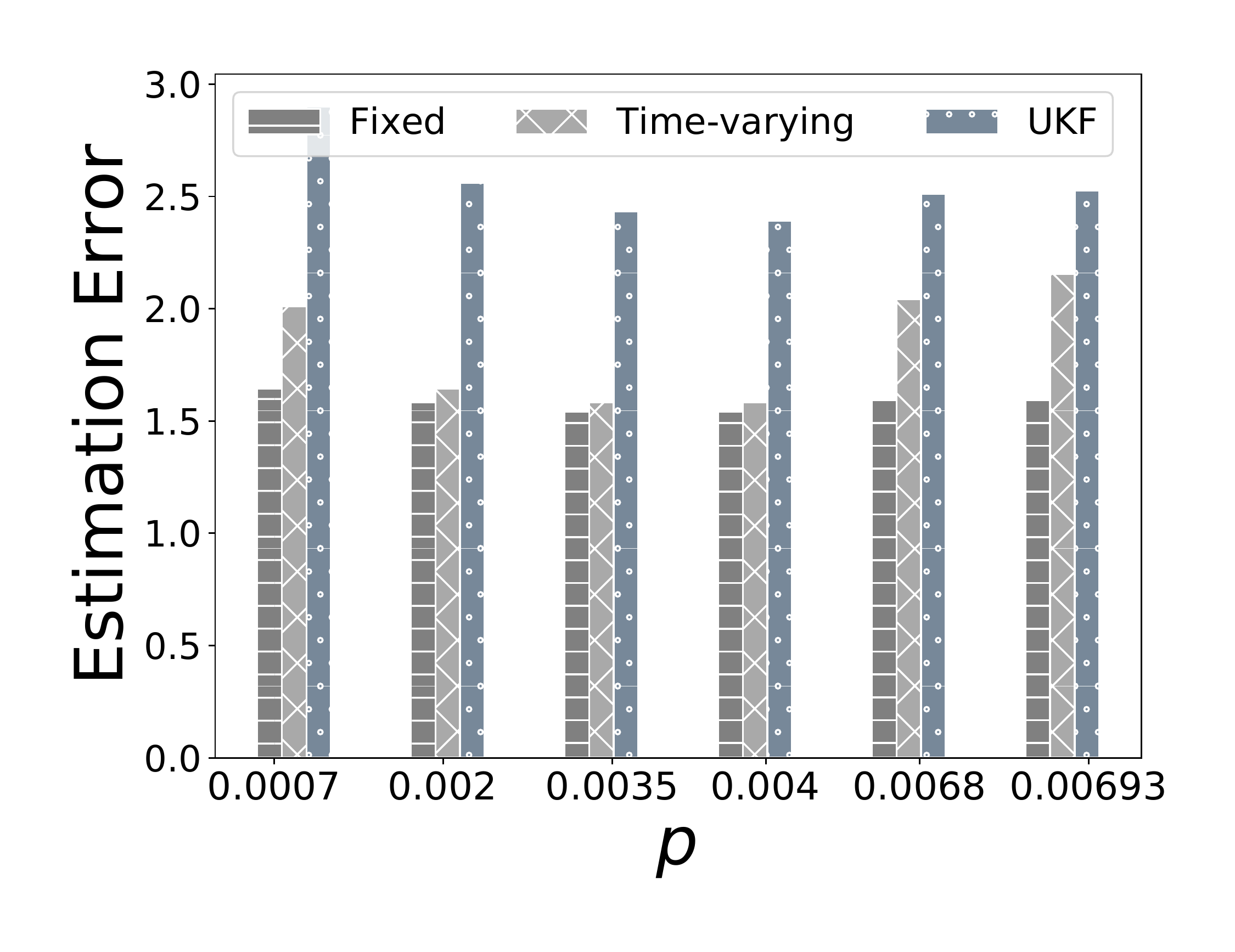}
			\label{fig:generalvsfix_nonlinear_0.0007}}
		\caption{\small The evaluation setup is akin to that for Figure~\ref{fig:TimeVaryingNtwkTrain}. We set $q=0.007$. (a) Linear system (b) Nonlinear System. Again, the LAA model trained over a time-varying network performs very well.}	
    	\label{fig:TimeVaryingNtwkTrain_0.007}		
	\end{center}
	\vspace{-0.125in}
\end{figure}

 To have to train LAA separately for fixed network settings would make it impractical. Often, the network parameters aren't known and change over time. Figures~\ref{fig:TimeVaryingNtwkTrain} and~\ref{fig:TimeVaryingNtwkTrain_0.007} compare the testing performance of LAA trained over a time-varying network with the testing performance of LAA trained for the same fixed network parameters that are used for testing. \emph{Figures~\ref{fig:generalvsfix_linear_0.3}, ~\ref{fig:generalvsfix_nonlinear_0.5},~\ref{fig:generalvsfix_linear_0.0007},~\ref{fig:generalvsfix_nonlinear_0.0007} show that the testing performance of the model trained using a time-varying network is in fact almost as good as that trained for the fixed network setting. There is no benefit in training different LAA models for different fixed network settings.} For Figure~\ref{fig:generalvsfix_linear_0.3}, the testing was done for $q=0.3$ and a linear system. In Figure~\ref{fig:generalvsfix_nonlinear_0.5}, the chosen system is the nonlinear cartpole and $q=0.5$. Figure~\ref{fig:TimeVaryingNtwkTrain_0.007} shows the estimation error for very small values of $p$ and $q$. The LAA model trained over a time-varying network is seen to work well over a wide range of $p$ and $q$.


\section{Conclusion}
\label{sec:conclusions}
We proposed a deep neural network model to learn estimates of state measurements in a model-free setting using intermittent and aged state measurements received over the network. We demonstrated its efficacy by comparing with the baselines of the TVKF and the UKF for different network settings and example linear and nonlinear systems. Our model performed well even in the presence of noisy age estimates and doesn't need to be trained separately for different network configurations, making it more suitable for real world settings. We showed that the age of updates as an input results in smaller estimation error. Also, all estimators see smaller errors when the utilization of the network minimizes the average age of updates at the estimator.

\bibliographystyle{IEEEtran}
\bibliography{literature}

\end{document}